\documentclass[11pt]{article}
\usepackage{acl}

\usepackage{fontspec}
\usepackage{xeCJK}
\IfFontExistsTF{SimSun}{\setCJKmainfont[ItalicFont=SimSun]{SimSun}}{\setCJKmainfont{gbsn00lp.ttf}}
\IfFontExistsTF{SimHei}{\setCJKsansfont{SimHei}}{\setCJKsansfont{gkai00mp.ttf}}
\IfFontExistsTF{SimSun}{\setCJKmonofont{SimSun}}{\setCJKmonofont{gbsn00lp.ttf}}

\usepackage{amsmath}
\usepackage{multirow}
\usepackage{url}
\usepackage{color}
\usepackage{graphicx}
\usepackage{booktabs}
\usepackage{array}
\newcommand{\modelname}{Qwen3.5-27B-Q4\_K\_M}
\makeatletter
\def\ps@pprintTitle{%
  \let\@oddhead\@empty
  \let\@evenhead\@empty
  \def\@oddfoot{}%
  \let\@evenfoot\@oddfoot
}
\makeatother
\begin{document}

\title{Operationalizing Linguistic Methods through Prompt-Engineering Skills: An Automatic Chinese Web Neologism Detection Pipeline}

\author{Yufeng Wu \\
  City University of Hong Kong \\
  \texttt{yufenwu2-c@my.cityu.edu.hk}
  \And
  Meichun Liu \\
  City University of Hong Kong}
\maketitle

\begin{abstract}
We present a method for automatic Chinese web neologism
detection that operationalizes traditional linguistic
identification principles as prompt-engineering skills.
The method has four stages: tokenizer-independent character
n-gram candidate generation; dictionary anchoring with a
Pointwise Mutual Information pre-filter; a well-formedness
skill based on Chinese word-formation principles; and a
combined rule and three-way classification skill that
distinguishes \textsc{neologism}, \textsc{entity}, and
\textsc{none}. Applied to the BAAI CCI 3.0 corpus (267M
documents), the method produces 226{,}959 classified
candidates including 4{,}853 labeled \textsc{neologisms}.
To evaluate the method, we develop a per-stage conditional
recall decomposition in which the pipeline's strict recall
factors mathematically into the product of stage conditional
recalls. Applied to \citet{houmin2023} (4{,}199 entries), the
decomposition exposes Stage 1 candidate coverage and Stage
4B LLM semantic judgment as the two bottlenecks
($R{=}41.5\%$ and $60.0\%$ respectively), while
intermediate stages are near-lossless. A length-stratified
analysis further reveals that the structural well-formedness
skill is length-invariant ($\geq{}96.9\%$) whereas the
semantic novelty-classification skill is length-dependent
($65.6\%/59.0\%/44.1\%$ across 2/3/4-character candidates),
mapping a current boundary of skill-based linguistic
operationalization. We release the method, pipeline outputs,
and evaluation protocol as public resources.
\end{abstract}

\section{Introduction}
\label{sec:intro}

Chinese web neologisms, new words and expressions emerging
from online language use, pose a persistent challenge for
computational lexicography. Their identification is
linguistically rich: traditional Chinese linguistics
provides well-developed accounts of word-formation
principles and of the boundary between novel coinages and
pre-existing lexical inventory. However, these accounts
have largely remained \emph{descriptive}, used by
lexicographers and linguists to manually identify new
words rather than to automatically detect them. Bridging
this gap, turning descriptive linguistic knowledge into
automatic detection procedures, has been a long-standing
goal of Chinese NLP.

Recent advances in large language models (LLMs) raise the
possibility that linguistic identification knowledge can
be operationalized through prompts. Yet writing a prompt
that says ``judge whether this is a new word'' is not
the same as encoding linguistic identification principles
in a reproducible form. In this paper we work within
current prompt-engineering practice and treat each
LLM-based classification stage as a \emph{skill}
\cite{anthropic2024skills}: a structured LLM-task interface
combining (a) explicit rule specifications grounded in
linguistic principles, (b) representative positive and
negative examples, and (c) format constraints on the
model output. The contribution of this paper is not to
introduce the notion of skill, which is part of
established prompt-engineering practice, but to
investigate how reliably linguistic identification
knowledge can be operationalized through this paradigm,
and where its boundaries lie.

We develop a four-stage pipeline for Chinese web neologism
detection from the BAAI CCI 3.0 corpus (267M documents).
The pipeline narrows from tokenizer-independent character
n-gram candidates, through dictionary-anchored and
PMI-based pre-filtering, to two LLM skills: a structural
well-formedness skill and a three-way semantic
classification skill, plus a rule-based fragment filter.
The final output is 226{,}959 classified candidates with
4{,}853 labeled \textsc{neologisms}, released as a public
resource. To evaluate the pipeline, we depart from
single-number recall reporting and develop a
\emph{per-stage conditional recall decomposition}: for
each entry in the \citet{houmin2023} reference dictionary
(4{,}199 entries), we trace its survival across all
five pipeline stages, computing the conditional recall
of each. The pipeline's aggregate strict recall factors
mathematically into the product of stage conditional
recalls, allowing diagnostic comparison at the
sub-pipeline level. We additionally decompose recall
by candidate character length to reveal length-dependent
patterns.

The method's diagnostic value is demonstrated by two
findings on the Hou (2023) reference set. First, the per-stage
decomposition exposes a bimodal bottleneck structure:
Stage 1 candidate coverage and Stage 4B LLM semantic
classification together account for 83.7\%\footnote{This is the joint figure (2,455+470)/3,495 = 83.69\%; the separately rounded component shares (70.2\% and 13.4\%) sum to 83.6\%.} of false
negatives, while intermediate stages are near-lossless.
Second, length-stratified diagnosis reveals a dissociation
between two LLM skills running on the same model
(\modelname{}): the structural well-formedness skill
(Stage~3) is length-invariant, retaining 96.86\%, 97.97\%,
and 100.00\% of 2-, 3-, and 4-character Hou (2023) entries
respectively, while the semantic three-way classification
skill (Stage~4B) is length-dependent, retaining only
65.62\%, 59.01\%, and 44.14\%. The same LLM, with
comparable skill design effort, succeeds at structural
word-formation judgment but degrades on semantic novelty
judgment as candidate length increases. This dissociation,
made visible by the method, maps a current boundary of
skill-based operationalization of linguistic
identification knowledge.

\paragraph{Contributions.} This paper makes four
methodological contributions: (1) a four-stage
tokenizer-independent method for Chinese web neologism
detection that operationalizes linguistic identification
principles as prompt-engineering skills, fully automatic
and applicable to any large Chinese corpus; (2) a
per-stage conditional recall evaluation protocol that
decomposes pipeline recall into multiplicative stage
factors, enabling diagnostic comparison at the sub-pipeline
level; (3) a public resource of 226{,}959 classified
candidates from CCI 3.0, including 4{,}853 labeled
\textsc{neologisms}; and (4) length-stratified diagnosis
that, applied to our method, reveals a structural-vs-
semantic skill boundary, observed for the first time at
this granularity in Chinese neologism detection.

\section{Related Work}
\label{sec:related}

Our work intersects two research strands: Chinese
neologism detection (traditional and LLM-based) and
prompt-engineering methodology with per-stage pipeline
evaluation. We review each in turn.

\subsection{Chinese Neologism Detection and LLM-based Lexical Approaches}

Chinese linguistic research has produced extensive
descriptive accounts of neologism formation and
identification. \citet{wu2000} and related work establish
core word-formation principles, modification
(\emph{piānzhèng} 偏正), verb-object (\emph{dòngbīn}
动宾), parallel (\emph{bìngliè} 并列), and affixation,
that govern productive new-word creation in Chinese.
\citet{ji2006} develop character-position rules for
filtering ill-formed Chinese n-gram fragments in the context
of terminology extraction, providing the linguistic basis for
our Stage~4A rule chain. \citet{feng2004} address the lexical boundary problem: which character sequences should be
treated as new words versus pre-existing inventory.
Computational approaches to Chinese new word detection
have evolved from rule-and-statistics hybrids to neural
methods over the past two decades. Early work combined
statistical sequence modeling with new-word boundary
identification \cite{peng2004crf}. Pretrained Chinese representations and transfer-learning segmentation models also improved lexical boundary modeling in adjacent Chinese NLP tasks \cite{sun2021chinesebert,huang2020multicriteria}. More recent task-specific
work applies supervised or semi-supervised detection to
downstream applications such as sentiment analysis
\cite{huang2014sentiment}. Domain-specific extensions target
social media and microblog vernacular, where neologism
density is highest. These approaches improve on rule-based
recall but typically evaluate against small annotated test
sets rather than against a comprehensive reference dictionary,
and they generally do not separate structural well-formedness
from semantic novelty, a distinction we make explicit through
our two-skill design.

These accounts are linguistically rich but remain
descriptive: they specify what constitutes a neologism
without specifying an executable detection procedure.
Our work takes these accounts as input and
operationalizes them as LLM skills (Stages 3, 4B)
plus rule-based and statistical filters (Stages 1,
2, 4A).

Building on these traditional accounts, recent LLM-based
approaches have begun automating neologism detection.

Recent work has applied LLMs to neologism and novel
lexeme detection. Most directly comparable is \citet{rossini2026}, who develop an English neologism
detection pipeline using a multi-model LLM ensemble
(Qwen 72B + Llama 70B + Mistral 123B + Claude Haiku
verification) and evaluate on a 53-entry reference
list. They report 37.7\% strict recall and a 58.7\%
true-novelty rate among classified candidates. \citet{rossini2026} use a four-class scheme (neologism, entity, foreign, none); we adopt three of these---neologism, entity, and none---and omit the foreign class, which targets English-language social-media text and is not pertinent to our Chinese setting. The 59.97\% strict
conditional recall of our Stage~4B closely matches
Rossini and van der Plas's 58.7\% across language and model
configuration, suggesting a cross-lingual ceiling for
current LLM-based neologism classification. For Chinese specifically, recent work has applied LLMs
to lexicon expansion and new-term identification, often
through prompt engineering or fine-tuning; however, we found
no high-confidence prior work that evaluates Chinese
neologism pipelines through per-stage conditional recall.
These applications mostly target social media monitoring,
trend tracking, or domain vocabulary discovery rather than
systematic pipeline evaluation against a comprehensive
reference dictionary. To our knowledge, no prior work
diagnoses Chinese neologism pipelines through per-stage
conditional recall against a reference list as comprehensive
as \citet{houmin2023}.

\subsection{Prompt Engineering and Per-stage Pipeline Evaluation}

A growing literature characterizes how task instructions
shape LLM behavior \citep{sahoo2024promptsurvey,schulhoff2024promptreport}. \citet{brown2020gpt3} establish
few-shot prompting as a primary mechanism for
in-context task specification. \citet{wei2022cot} show
that chain-of-thought reasoning improves multi-step
performance. Beyond ad-hoc prompts, structured
LLM-task interfaces, variously termed \emph{skills}
\cite{anthropic2024skills}, \emph{instructions}, or
\emph{task specifications}, combine rule descriptions,
representative examples, and format constraints to
make LLM behavior more reproducible and auditable. Recent work has also developed structured output mechanisms for LLMs, including JSON-mode decoding, function calling, and constrained generation \citep{outlines2023,geng2025jsonschemabench,wang2025slot}, which underlies the format-constraint component of our skill design.
More broadly, the rapid growth of LLM evaluation benchmarks motivates reporting diagnostic stage-level behavior rather than a single aggregate score \cite{ni2025llmbenchmarks}. Our work uses this paradigm: each LLM stage in the
pipeline is a skill in this sense, with a fixed rule
specification, a curated example set, and explicit
output constraints. We do not introduce the notion of
skill; our contribution is to investigate how reliably
linguistic identification knowledge can be
operationalized through it.

This structured-task view connects naturally to per-stage
evaluation, because each prompt-defined component can be
audited as a separate pipeline stage.

Pipeline error attribution is a long-standing concern
in NLP. Component-wise evaluation has been used to diagnose machine translation pipelines \cite{vilar2006} and parsing pipelines \cite{smith-eisner-2008}, among others. Per-stage conditional recall, as we
develop it here, applies this diagnostic philosophy
specifically to neologism detection: by tracing the
fate of each reference-list entry through all pipeline
stages and computing each stage's input-to-output
retention, we make the multiplicative structure of
aggregate recall explicit. This complements aggregate
recall reporting, allowing methodological comparisons
at the stage level rather than at the pipeline level.

\section{Methods}
\label{sec:methods}

This section describes the four-stage pipeline. We first
give an overview (§\ref{subsec:overview-method}) and
then describe each stage in turn: candidate generation
(§\ref{subsec:stage1}), vocabulary anchoring and
pre-filtering (§\ref{subsec:stage2}), LLM
well-formedness skill (§\ref{subsec:stage3}), three-way
classification (§\ref{subsec:stage4}), output
construction (§\ref{subsec:output}), and evaluation
methodology (§\ref{subsec:eval}).

\subsection{Overview}
\label{subsec:overview-method}

We propose a four-stage pipeline for detecting Chinese web neologisms from large-scale corpora. The pipeline progressively narrows from raw character n-gram statistics through dictionary-anchored filtering, LLM-based well-formedness classification, and a combined rule-based and LLM-based three-way classification. We use a single open-source LLM (\modelname{}) throughout \cite{qwen35_27b_2026}. We organize the LLM components of the pipeline as prompt-engineering skills \citep{anthropic2024skills}: structured LLM-task interfaces consisting of (a) explicit rule specifications grounded in linguistic principles, (b) representative positive and negative examples, and (c) format constraints on the model output. This framing allows us to operationalize traditionally descriptive linguistic identification knowledge into reproducible, model-substitutable LLM tasks. We deploy two skills in the pipeline: a structural well-formedness skill (Stage 3) and a three-way semantic classification skill (Stage 4B).

\begin{figure*}[!htbp]
\centering
\includegraphics[width=\textwidth]{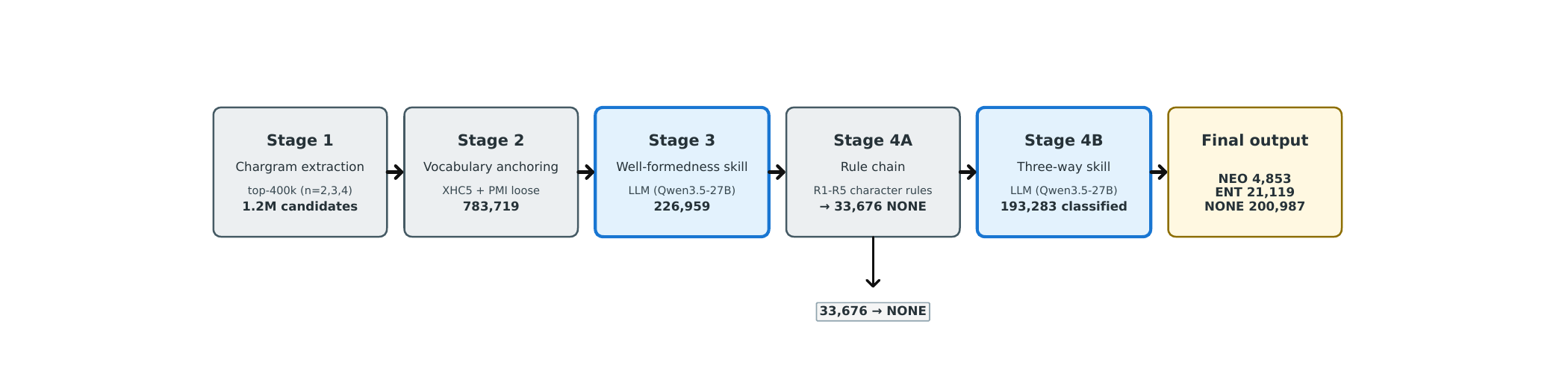}
\caption{Four-stage pipeline for Chinese web neologism detection. Stage 3 and Stage 4B use LLM-based prompt-engineering skills (highlighted in blue).}
\label{fig:pipeline}
\end{figure*}

\subsection{Corpus and Candidate Generation (Stage 1)}
\label{subsec:stage1}

We use the BAAI CCI 3.0 corpus, a curated Chinese web corpus of 267,655,159 documents. To avoid the lexical bias inherent in tokenizer-dependent pipelines, where a tokenizer's built-in vocabulary pre-determines which character sequences can become word candidates, we extract candidates directly via character n-gram counting without any word-level tokenization. We extract n-grams of lengths 2, 3, and 4, retaining the top 400,000 by raw frequency for each length, yielding a candidate pool of 1.2 million entries.

\subsection{Vocabulary Anchoring and Statistical 
Pre-filtering (Stage 2)} \label{subsec:stage2}

We anchor against the Modern Chinese Dictionary, 5th edition
(XHC5, 2005): candidates appearing in XHC5 are removed,
isolating items absent from the pre-2005 standard lexicon.
Wikipedia titles (965K entries) are recorded as an
\texttt{in\_wiki} flag for downstream reference but not used
to exclude candidates, since many web neologisms
(\emph{nèijuǎn} 内卷 'involution', \emph{fáněrsài} 凡尔赛
'humble-bragging') have Wikipedia entries. To control
downstream cost, we apply a Pointwise Mutual Information
(PMI) pre-filter: PMI ${>}2$ for 2-/3-character candidates,
PMI$^2$ ${>}4$ for 4-character. We label this ``PMI loose''
and acknowledge it as a computational compromise; some
cohesively-weak neologisms with negative PMI (\emph{shèsǐ}
社死 'social death', \emph{dāzi} 搭子 'companion') are lost
here. After Stage 2: 783{,}719 candidates.

\subsection{LLM Well-formedness Skill (Stage 3)}
\label{subsec:stage3}

The first LLM skill (\modelname{}, via llama.cpp)
classifies each candidate as WELL\_FORMED or FRAGMENT.
The skill specifies Chinese word-formation
principles, including modification (\emph{piānzhèng} 偏正),
verb-object (\emph{dòngbīn} 动宾), parallel
(\emph{bìngliè} 并列), quadrisyllabic templates, and
affixation, as the structural criterion, with representative
examples and a single-token output format. Candidates
satisfying word-formation rules are retained even when
unrecognized by the LLM, ensuring structural rather than
memory-based judgment. After Stage 3: 226{,}959
WELL\_FORMED candidates.

\subsection{Three-way Classification (Stage 4)}
\label{subsec:stage4}

Stage 4 classifies each WELL\_FORMED candidate into NEOLOGISM, ENTITY, or NONE through two components: a rule-based pre-filter (4A) and an LLM-based classifier (4B).

\paragraph{Stage 4A: Rule-based Pre-filtering.}

We apply five independent character-position rules (Table~\ref{tab:rule-config}), each targeting a distinct surface pattern characteristic of NONE candidates. Each rule examines a single character position (initial or final) and matches against a closed character list derived from the XHC5 character-frequency top-300 band.

\begin{table*}[t]
\centering
\small
\caption{Stage 4A rule chain configuration}
\label{tab:rule-config}
\resizebox{\textwidth}{!}{%
\begin{tabular}{l l p{0.5\linewidth} r}
\toprule
Rule & Position & Pattern & List size \\
\midrule
R1 & initial & function-word prefix & 15 \\
R2 & final & function-word / locative / temporal suffix & 19 \\
R3 & initial & negation / adverb / preposition prefix & 22 \\
R4 & initial & determiner prefix & 6 \\
R5 & initial & sentence-cut verb head & 4 \\
\bottomrule
\end{tabular}
}
\end{table*}

The R2 and R3 lists were extended with locative-temporal characters (中, 上, 下, 里, 内, 外, 前, 后, 间, 时) and prepositional characters (在, 于, 把, 对, 向, 为, 以, 自, 由) based on prior work on Chinese garbage-string filtering \cite{ji2006}. All character lists are validated against an internal target list of representative web neologisms.

Stage 4A filters 33{,}676 candidates (14.84\% of input) to NONE; the remaining 193{,}283 candidates pass to Stage 4B.

\paragraph{Stage 4B: Three-way Classification Skill.}

The same \modelname{} applies a three-way classification
skill to the 193{,}283 remaining candidates.
The skill specifies class definitions, representative
examples (12 NEOLOGISM, 10 ENTITY, 21 NONE), and seven
decision rules: derivative forms remain NEOLOGISM; the
\texttt{in\_wiki} flag is reference-only; existing fixed
expressions and domain terminology remain NONE; uncertainty
defaults to NONE. The LLM runs at temperature 0, single-pass.
Of 193{,}283 candidates, parsing failed for 16 (0.008\%) and
were conservatively assigned NONE. Output: 4{,}853
NEOLOGISM, 21{,}119 ENTITY, 167{,}311 NONE.

\subsection{Final Output and Source Annotation}
\label{subsec:output}

Stage 4A and Stage 4B outputs are merged into the final candidate set, with each candidate annotated by its decision source (R1--R5 or LLM) to enable downstream interpretability and error analysis.

\subsection{Evaluation Methodology}
\label{subsec:eval}

We evaluate against the New Chinese Words dictionary of
\citet{houmin2023}, normalized to 4{,}199 unique entries (coverage
2000--2020). For each entry, we trace its survival across
all five pipeline stages and compute the conditional recall
$R_k$ of stage $k$ as the fraction of Hou (2023) entries
entering $k$ that survive. The pipeline's overall strict
recall equals the product $R_1 \times R_2 \times R_3 \times
R_{4A} \times R_{4B}$. We additionally decompose recall by
candidate character length (2/3/4-char) to expose
length-dependent bottlenecks. This decomposition provides
a diagnostic framework rather than a single performance
number.

\section{Results}
\label{sec:results}

This section reports pipeline output statistics
(§\ref{subsec:dataflow}) and the evaluation against
\citet{houmin2023}: per-stage conditional recall
(§\ref{subsec:per-stage}), recall decomposition by
candidate length (§\ref{subsec:by-length}), and the
internal fate of Hou (2023) entries reaching Stage 4
(§\ref{subsec:internal}).

\subsection{Pipeline Data Flow}
\label{subsec:dataflow}

Table~\ref{tab:dataflow} shows candidate counts at each pipeline stage.

\begin{table*}[t]
\centering
\small
\caption{Stage-by-stage candidate counts}
\label{tab:dataflow}
\resizebox{\textwidth}{!}{%
\begin{tabular}{lp{0.56\linewidth}r}
\toprule
Stage & Operation & Output \\
\midrule
Stage 1 & character n-gram top-400k (lengths 2, 3, 4) & 1,200,000 \\
Stage 2 & XHC5 anchor + PMI loose pre-filter & 783,719 \\
Stage 3 & LLM well-formedness filter & 226,959 \\
Stage 4A & rule-based NONE filter & 33,676 $\rightarrow$ NONE \\
Stage 4B & LLM three-way classification & 193,283 \\
Final & merged output & 226,959 \\
\bottomrule
\end{tabular}
}
\end{table*}

The final classification distribution is: NEOLOGISM 4,853 (2.14\%), ENTITY 21,119 (9.30\%), NONE 200,987 (88.56\%). The conservative NEOLOGISM proportion reflects the design choice that uncertainty defaults to NONE.

\subsection{Per-stage Conditional Recall}
\label{subsec:per-stage}

Table~\ref{tab:per-stage} reports the per-stage conditional recall of \citet{houmin2023} entries through the pipeline.

\begin{table*}[t]
\centering
\small
\caption{Per-stage conditional recall against Hou (2023), n = 4,199}
\label{tab:per-stage}
\resizebox{\textwidth}{!}{%
\begin{tabular}{lrrr}
\toprule
Stage & Entering & Surviving & Recall [95\% Wilson CI] \\
\midrule
Stage 1: character n-gram extraction & 4,199 & 1,744 & 41.53 [40.05, 43.03] \\
Stage 2: XHC5 + PMI loose & 1,744 & 1,226 & 70.30 [68.11, 72.40] \\
Stage 3: LLM well-formedness skill & 1,226 & 1,198 & 97.72 [96.72, 98.42] \\
Stage 4A: rule pre-filter & 1,198 & 1,174 & 98.00 [97.04, 98.65] \\
Stage 4B: LLM 3-way skill (strict NEO) & 1,174 & 704 & 59.97 [57.14, 62.73] \\
Pipeline total & 4,199 & 704 & 16.77 \\
\bottomrule
\end{tabular}
}
\end{table*}

Multiplicative identity (verified to 12 decimal places):

$R_1 \times R_2 \times R_3 \times R_{4A} \times R_{4B}$ = 0.4153 × 0.7030 × 0.9772 × 0.9800 × 0.5997 = 0.16766 = pipeline strict recall.

Under the inclusive criterion (treating ENTITY as acceptable lexical innovation), Stage 4B conditional recall rises to 807 / 1,174 = 68.74\%, yielding inclusive pipeline recall of 19.22\%.

The decomposition reveals a bimodal pipeline: Stages 3 and 4A are near-lossless (R > 97\%), Stage 2 is moderate (R = 70\%), while Stages 1 and 4B are the two major recall bottlenecks (R = 41.5\% and 60.0\% respectively). Detailed false negative distribution by stage and component is provided in Appendix~\ref{app:fn-dist}.

\begin{figure}[!htbp]
\centering
\includegraphics[width=0.9\linewidth]{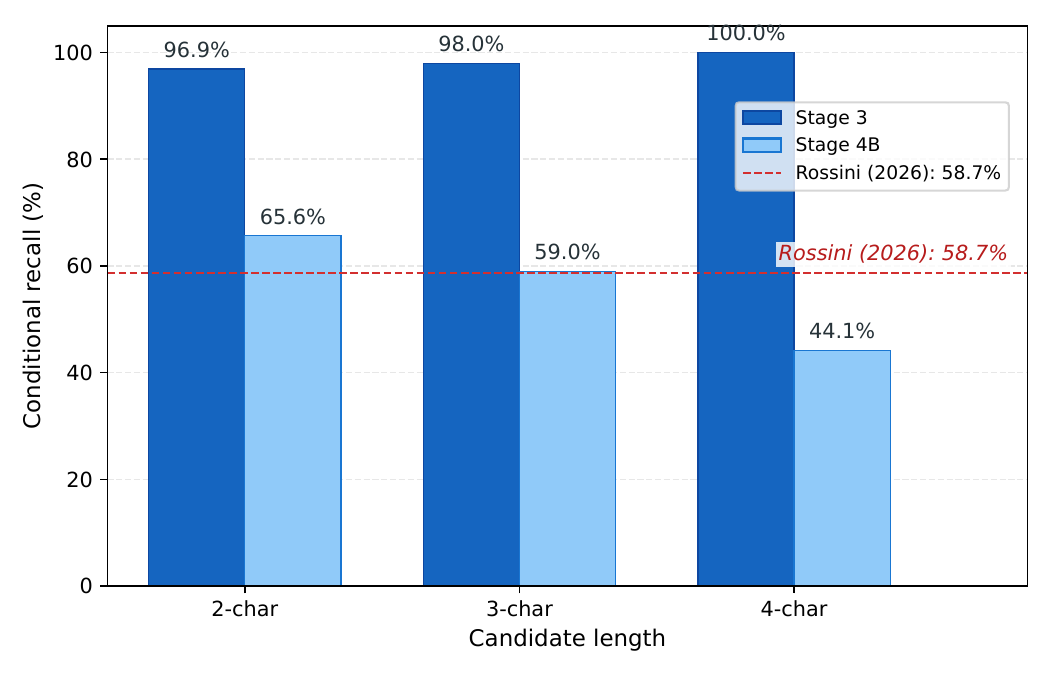}
\caption{Conditional recall comparison between Stage 3 well-formedness skill (length-invariant) and Stage 4B three-way classification skill (length-dependent), across 2/3/4-character Hou (2023) entries. The dashed line indicates the 58.7\% true-novelty rate reported by Rossini and van der Plas (2026) for English neologism detection.}
\label{fig:skill-length}
\end{figure}

\subsection{Recall Decomposition by Candidate Length}
\label{subsec:by-length}

The pipeline's per-stage conditional recall varies systematically with candidate length. Appendix Table~\ref{tab:by-length} reports the same per-stage decomposition broken down into 2-character, 3-character, and 4-character Hou (2023) entries (constituting 50.3\%, 29.6\%, and 15.6\% of the reference list respectively).

Note: 5+-character Hou (2023) entries (n=130) are excluded from candidate generation at Stage 1 by design (character n-gram lengths 2--4 only); 1-character entries (n=61) are similarly out of scope.

The by-length decomposition reveals three structural observations absent from the aggregate report:

2-character new words are bottlenecked at Stages 1 and 2. The Stage 1 top-400k threshold misses 43\% of 2-char Hou (2023) entries, and Stage 2's PMI loose filter further removes 42\% of survivors. PMI as a cohesion measure is poorly suited to 2-character new words because many genuine 2-char neologisms, particularly those derived from sound symbolism, abbreviation, or weak co-occurrence patterns (e.g., \emph{shèsǐ} 社死~'social death', \emph{dāzi} 搭子~'companion'), exhibit low PMI by construction.

3-character new words are bottlenecked overwhelmingly at Stage 1. Only 24.9\% of 3-char Hou (2023) entries reach Stage 2; once they pass, Stages 2--4A retain >95\% of them. The 3-char Stage 1 bottleneck reflects that productive 3-character neologisms (X化, X性, X感, V-N constructions) tend to occupy the moderate-frequency range that falls just outside top-400k coverage.

4-character new words present a distinct pattern: they survive Stages 1--4A relatively well (Stage 3 retains 100\%, Stage 4A retains 96.5\%) but suffer the most severe drop at Stage 4B (44.1\% strict recall). This reflects the fundamental ambiguity at length 4 between novel coinages and the pre-existing inventory of four-character fixed expressions (chengyu, idiomatic compounds, technical terms).

The contrast between Stage 3 (length-invariant: 96.86/97.97/100.00\%) and Stage 4B (length-dependent: 65.62/59.01/44.14\%) is methodologically informative. The Stage~4B decline is significant (Cochran--Armitage trend test, $z=-5.54$, $p=2.97{\times}10^{-8}$), as is the stage-by-length interaction between Stage~3 and Stage~4B (empirical-logit interaction, $z=-2.67$, $p=0.00765$). The same \modelname{} model is reliable when asked to perform structural word-formation judgment but degrades under semantic neologism judgment, with degradation correlating with length-specific lexical complexity.

\subsection{Stage 4 Internal Diagnosis}
\label{subsec:internal}

Among the 1,198 Hou (2023) entries that survive to Stage 4, Appendix Table~\ref{tab:stage4-diag} shows the internal classification fate.

Among the 24 Stage 4A rule false rejections, affected entries include productive prefixed terms such as \emph{yǐfáng yǎnglǎo} 以房养老~'house-for-pension' (R1 on initial 以), \emph{zìjiàyóu} 自驾游~'self-drive tourism' (R3 on initial 自), and \emph{zàixiàn jiàoyù} 在线教育~'online education' (R3 on initial 在). These are systematic cases where productive Chinese word-formation overlaps with rule detection patterns.

Among the 103 Stage 4B ENTITY classifications, entries are predominantly composite terms denoting new policy regimes, products, or social phenomena (e.g., \emph{Bāxiàng Guīdìng} 八项规定~'Eight-Point Regulation', \emph{Pīnduōduō} 拼多多~'e-commerce platform', \emph{Hùgǎngtōng} 沪港通~'Shanghai-Hong Kong Stock Connect', \emph{Huǒjiànjūn} 火箭军~'PLA Rocket Force'). Each is simultaneously a proper-noun-like designation and a productively-coined new term, a category boundary that \citet{rossini2026} describe as ``inherently fuzzy.''

Among the 367 Stage 4B NONE misclassifications, the genuine LLM errors, examples include \emph{xuānguàn} 宣贯~'promulgation and implementation', \emph{xiǎo xīnsi} 小心思~'little intentions', \emph{tuìgǔ} 退股~'stock withdrawal', \emph{chūnzhāo} 春招~'spring recruitment', \emph{gōngzūfáng} 公租房~'public rental housing', \emph{chùkòngpíng} 触控屏~'touchscreen', \emph{miànqiān} 面签~'in-person signing'. These represent the residual semantic ambiguity that remains after rule-based and dictionary-based pre-filtering.

\section{Discussion}
\label{sec:discussion}

This section interprets the evaluation results. We
first discuss what the pipeline's aggregate recall
does and does not say
(§\ref{subsec:what-recall-says}), then examine the two
bottleneck stages
(§\ref{subsec:two-bottlenecks}), then turn to the
length-dependent structure of pipeline errors
(§\ref{subsec:length-dep}) as our principal finding,
and finally draw methodological implications for
future pipeline design (§\ref{subsec:implications}).
Limitations are discussed separately in
Section~\ref{sec:limitations}.

\subsection{What This Pipeline's Recall Tells Us}
\label{subsec:what-recall-says}

The pipeline's strict recall against \citet{houmin2023} is 16.77\%, lower than the 37.7\% reported by \citet{rossini2026} for English. However, direct comparison is not meaningful: Rossini and van der Plas's evaluation uses a curated 53-entry reference list, whereas \citet{houmin2023} contains 4,199 entries including a long tail of low-frequency expressions. More fundamentally, a single aggregate recall number conflates loss across five sequential filtering stages of very different fidelity.

The per-stage decomposition (Table~\ref{tab:per-stage}) is the principal contribution of our evaluation. It reveals three near-lossless stages: Stages 3, 4A, and (to a lesser extent) Stage 2 retain >70\% of their input each, with Stages 3 and 4A retaining >97\%. Combined recall loss across Stages 2--4A is only 16.3\% of total FN. It also reveals two bottleneck stages: Stage 1 retains only 41.5\% of Hou (2023) entries, and Stage 4B's strict NEOLOGISM classification retains only 60.0\% of its input. These two stages account for 83.7\% of all false negatives.

This decomposition reframes the evaluation: the question is no longer ``is our pipeline good or bad?'' but ``where in the pipeline does recall actually leak, and why?''

\subsection{Two Bottlenecks of Very Different Natures}
\label{subsec:two-bottlenecks}

The Stage 1 bottleneck is fundamentally about candidate pool coverage. Of the 4,199 Hou (2023) entries, 2,264 (54\%) fall below the top-400k frequency threshold in CCI 3.0; a separate 191 lie outside the 2--4-character candidate-generation scope. Stage~1 thus accounts for 2,455 (58.5\%) of all reference entries lost. This reflects a property of the new-word distribution itself: by definition, new words are not yet widespread, and many remain rare even in large web corpora. Recovering them requires either a larger candidate pool (which would multiply downstream LLM compute proportionally) or candidate-generation strategies orthogonal to raw frequency, such as productivity-based extraction or domain-specific corpus augmentation.

The Stage 4B bottleneck is about LLM semantic judgment of neologism status. The 59.97\% strict conditional recall is remarkably close to the 58.7\% true-novelty rate reported by \citet{rossini2026} for English neologism detection using a multi-model ensemble (Qwen 72B + Llama 70B + Mistral 123B + Claude Haiku verification). The cross-linguistic and cross-configuration convergence near 60\% suggests this number may reflect a current ceiling for LLM-based neologism classification rather than a Chinese-specific or single-model limitation.

The boundary cases driving Stage 4B errors are themselves theoretically ambiguous. V+N constructions (e.g., \emph{tāfáng} 塌房~'idol collapse' vs. \emph{xǐwǎn} 洗碗~'wash dishes') are character-pattern-identical to common verb-object expressions; ENTITY-NEOLOGISM dual designations (\emph{Pīnduōduō} 拼多多~'e-commerce platform', \emph{Hùgǎngtōng} 沪港通~'Shanghai-Hong Kong Stock Connect', \emph{Huǒjiànjūn} 火箭军~'PLA Rocket Force') resist clean categorical assignment; and domain terms (\emph{chūnzhāo} 春招~'spring recruitment', \emph{gāosòngzhuǎn} 高送转~'high stock dividend') sit at the disciplinary boundary of which fields' technical vocabularies count as ``general'' Chinese.

\subsection{The Length-Dependent Structure of Pipeline Errors}
\label{subsec:length-dep}

The by-length decomposition (Appendix Table~\ref{tab:by-length} and
Figure~\ref{fig:skill-length}) shows that recall bottlenecks
are length-specific rather than uniform: 2-character entries
are bottlenecked at Stages 1--2 (long-tail frequency and
low PMI of 2-char neologisms), 3-character entries primarily
at Stage~1 (top-400k coverage), and 4-character entries at
Stage~4B alone (44.1\% strict recall, the lowest at any
stage-length combination). The contrast between Stage 3
(length-invariant: 96.86/97.97/100.00\%) and Stage 4B
(length-dependent: 65.62/59.01/44.14\%) running on the same
\modelname{} is methodologically informative. The Stage~4B
length trend is significant ($z=-5.54$,
$p=2.97{\times}10^{-8}$), and the stage-by-length
interaction is also significant ($z=-2.67$, $p=0.00765$).
The model is
reliable at structural word-formation judgment but degrades
at semantic novelty judgment as length increases. We
summarize this as: a structural well-formedness skill is
length-invariant in its operationalization, whereas a
semantic novelty-classification skill is length-dependent
and degrades sharply on longer candidates. This
dissociation, observed through our method's per-stage
decomposition, maps the current boundary of skill-based
operationalization of linguistic identification knowledge.

\subsection{Methodological Implications}
\label{subsec:implications}

Three implications follow. First, mid-stages (Stages 2--4A)
are not the bottleneck; investment in alternative dictionary
anchors, PMI thresholds, or well-formedness criteria cannot
improve aggregate recall by more than a few percentage
points. Second, Stage~1 expansion has a clear cost-benefit
calculus: doubling the candidate pool to top-1M would
$2.5{\times}$ downstream LLM compute but could recover an
estimated 1{,}000--1{,}500 additional Hou (2023) entries,
raising pipeline recall to roughly 36--38\%. Third, Stage~4B
improvement may require fundamentally different approaches:
multi-model ensembles did not exceed the ${\sim}60\%$
ceiling, suggesting retrieval-augmented LLM judgment,
neologism-specific fine-tuning, or hybrid human-in-the-loop
designs as productive alternatives. More broadly, this
points to an open question for skill-based
operationalization: which categories of linguistic judgment
can be reliably encoded as LLM skills with current
prompt-engineering practice, and which require alternative
paradigms.

Per-stage and per-length conditional recall should
complement aggregate recall as evaluation criteria for
pipeline-style methods.

\section{Limitations}
\label{sec:limitations}

We acknowledge several limitations of this work.

\paragraph{Statistical pre-filter design cost.}
The PMI loose pre-filter in Stage 2 is a computational
compromise that loses cohesively-weak neologisms with
negative PMI, including \emph{shèsǐ} 社死 'social
death' and \emph{dāzi} 搭子 'companion'. This is
documented in our per-length analysis (Table~5)
but not corrected; alternative cohesion measures for
2-character Chinese neologisms remain an open problem.
Similarly, Stage 1's top-400k frequency threshold
excludes 54\% of Hou (2023) entries that fall in the
low-frequency long tail, a fundamental property of
neologism distributions in web corpora.

\paragraph{Dictionary anchor coverage.}
The XHC5 anchor (2005) is a single dictionary snapshot.
Some domain-specific lexical items predating 2005 but
not included in XHC5 may be incorrectly accepted as
candidates and surface in our final output. Conversely,
some Hou (2023) entries that XHC5 happens to contain are
excluded at Stage 2 (42 entries in our trace).

\paragraph{Reference list biases.}
The \citet{houmin2023} reference list, while the most
comprehensive Chinese neologism dictionary publicly
available, has its own selection biases: it favors
words attested in mainstream media, may underrepresent
purely online vernacular, and reflects editorial
choices about which expressions qualify as
``new words.'' Our recall numbers are therefore
relative to this specific reference frame and may
not generalize to all neologism definitions.

\paragraph{No precision evaluation.}
Our evaluation reports recall against \citet{houmin2023}
but does not include precision measurement against
human annotation of our pipeline's 4{,}853
\textsc{neologism} outputs. Whether our pipeline's
positive predictions are themselves neologisms, rather than
false positives, remains to be measured
in future work, ideally through stratified human
annotation with inter-annotator agreement.

\paragraph{Single LLM and reproducibility.}
We use a single open-source model (\modelname{})
at temperature 0 for both LLM skills. While temperature
0 reduces stochasticity, the specific weight
quantization and inference engine (llama.cpp) may
affect reproducibility for readers using different
runtime configurations. Our released outputs are
deterministic relative to our exact runtime setup.

\section{Conclusion}
\label{sec:conclusion}

We presented a method for automatic Chinese web neologism
detection that operationalizes traditional linguistic
identification principles as prompt-engineering skills,
combining tokenizer-independent character n-gram candidate
generation, dictionary anchoring, statistical pre-filtering,
a well-formedness skill, and a rule-plus-classification
stage. Applied to CCI 3.0 and evaluated against \citet{houmin2023}, the method enables per-stage conditional recall
decomposition that exposes bottleneck structure rather than
collapsing it into a single number. The decomposition
reveals two bottleneck stages (candidate coverage and LLM
semantic classification) and a length-dependent dissociation
between structural and semantic LLM skills running on the
same model, a boundary made visible by the method's
multiplicative recall structure. We release 226{,}959
classified candidates, the rule specifications, the
skill prompts, and the per-stage evaluation protocol as
public resources. Future work should explore alternative
candidate generation strategies beyond raw frequency, and
LLM-based semantic novelty judgment paradigms beyond
conventional skill design.
\bibliography{references}

@misc{anthropic2024skills,
  author = {{Anthropic}},
  title = {Claude Skills Documentation},
  year = {2024},
  url = {https://docs.claude.com/en/docs/agents-and-tools/skills/overview},
  note = {Accessed 2026-06-03}
}

@book{houmin2023,
  author = {Hou, Min},
  title  = {A Dictionary of New Chinese Words (2000--2020)},
  year   = {2023},
  publisher = {The Commercial Press},
  address = {Beijing},
  isbn = {978-7-100-21777-4},
  note = {In Chinese}
}

@misc{rossini2026,
  author = {Rossini, Diego and van der Plas, Lonneke},
  title = {From 124 Million Tokens to 1,021 Neologisms: A Large-Scale Pipeline for Automatic Neologism Detection},
  year = {2026},
  eprint = {2605.06426},
  archivePrefix = {arXiv},
  url = {https://arxiv.org/abs/2605.06426},
  doi = {10.48550/arXiv.2605.06426}
}

@inproceedings{wu2000,
  author = {Wu, Andi and Jiang, Zixin},
  title = {Statistically-Enhanced New Word Identification in a Rule-Based Chinese System},
  booktitle = {Proceedings of the Second Chinese Language Processing Workshop},
  pages = {46--51},
  year = {2000},
  url = {https://aclanthology.org/W00-1206/}
}

@article{feng2004,
  author = {Feng, Haodi and Chen, Kang and Deng, Xiaotie and Zheng, Weimin},
  title = {Accessor Variety Criteria for Chinese Word Extraction},
  journal = {Computational Linguistics},
  volume = {30},
  number = {1},
  pages = {75--93},
  year = {2004},
  doi = {10.1162/089120104773633376}
}

@inproceedings{ji2006,
  author = {Ji, Luning and Lu, Qin and Li, Wenjie and Chen, Yirong},
  title = {A Comparative Study of the Effect of Word Segmentation On Chinese Terminology Extraction},
  booktitle = {Proceedings of the 20th Pacific Asia Conference on Language, Information and Computation},
  pages = {101--108},
  year = {2006},
  address = {Huazhong Normal University, Wuhan, China},
  publisher = {Tsinghua University Press},
  url = {https://aclanthology.org/Y06-1014/}
}

@inproceedings{peng2004crf,
  author = {Peng, Fuchun and Feng, Fangfang and McCallum, Andrew},
  title = {Chinese Segmentation and New Word Detection using Conditional Random Fields},
  booktitle = {Proceedings of COLING 2004},
  year = {2004},
  url = {https://aclanthology.org/C04-1081/}
}

@inproceedings{huang2014sentiment,
  author = {Huang, Minlie and Ye, Bo and Wang, Yichen and Chen, Haiqiang and Cheng, Junjun and Zhu, Xiaoyan},
  title = {New Word Detection for Sentiment Analysis},
  booktitle = {Proceedings of ACL 2014},
  pages = {531--541},
  year = {2014},
  url = {https://aclanthology.org/P14-2052/},
  doi = {10.3115/v1/P14-2052}
}

@inproceedings{brown2020gpt3,
  author = {Brown, Tom B. and Mann, Benjamin and Ryder, Nick and Subbiah, Melanie and Kaplan, Jared and Dhariwal, Prafulla and Neelakantan, Arvind and Shyam, Pranav and Sastry, Girish and Askell, Amanda and Agarwal, Sandhini and Herbert-Voss, Ariel and Krueger, Gretchen and Henighan, Tom and Child, Rewon and Ramesh, Aditya and Ziegler, Daniel M. and Wu, Jeffrey and Winter, Clemens and Hesse, Christopher and Chen, Mark and Sigler, Eric and Litwin, Mateusz and Gray, Scott and Chess, Benjamin and Clark, Jack and Berner, Christopher and McCandlish, Sam and Radford, Alec and Sutskever, Ilya and Amodei, Dario},
  title = {Language Models are Few-Shot Learners},
  booktitle = {Advances in Neural Information Processing Systems},
  volume = {33},
  pages = {1877--1901},
  year = {2020},
  url = {https://arxiv.org/abs/2005.14165},
  doi = {10.48550/arXiv.2005.14165}
}

@inproceedings{wei2022cot,
  author = {Wei, Jason and Wang, Xuezhi and Schuurmans, Dale and Bosma, Maarten and Ichter, Brian and Xia, Fei and Chi, Ed and Le, Quoc and Zhou, Denny},
  title = {Chain-of-Thought Prompting Elicits Reasoning in Large Language Models},
  booktitle = {Advances in Neural Information Processing Systems},
  volume = {35},
  pages = {24824--24837},
  year = {2022},
  url = {https://proceedings.neurips.cc/paper_files/paper/2022/hash/9d5609613524ecf4f15af0f7b31abca4-Abstract-Conference.html}
}

@misc{outlines2023,
  author = {Willard, Brandon T. and Louf, R{\'e}mi},
  title = {Efficient Guided Generation for Large Language Models},
  year = {2023},
  eprint = {2307.09702},
  archivePrefix = {arXiv},
  url = {https://arxiv.org/abs/2307.09702}
}

@inproceedings{vilar2006,
  author = {Vilar, David and Xu, Jia and D'Haro, Luis Fernando and Ney, Hermann},
  title = {Error Analysis of Statistical Machine Translation Output},
  booktitle = {Proceedings of LREC 2006},
  pages = {697--702},
  year = {2006},
  url = {http://www.lrec-conf.org/proceedings/lrec2006/pdf/413_pdf.pdf}
}

@inproceedings{smith-eisner-2008,
  author = {Smith, David A. and Eisner, Jason},
  title = {Dependency Parsing by Belief Propagation},
  booktitle = {Proceedings of EMNLP 2008},
  pages = {145--156},
  year = {2008},
  url = {https://aclanthology.org/D08-1016/}
}

@misc{qwen35_27b_2026,
  author = {{Qwen Team}},
  title = {Qwen3.5-27B},
  year = {2026},
  url = {https://huggingface.co/Qwen/Qwen3.5-27B},
  note = {Hugging Face model card, model version Qwen/Qwen3.5-27B; accessed 2026-06-07}
}

@misc{schulhoff2024promptreport,
  author = {Schulhoff, Sander and Ilie, Michael and Balepur, Nishant and Kahadze, Konstantine and Liu, Amanda and Si, Chenglei and Li, Yinheng and Gupta, Aayush and Han, HyoJung and Schulhoff, Sevien and Dulepet, Pranav Sandeep and Vidyadhara, Saurav and Ki, Dayeon and Agrawal, Sweta and Pham, Chau and Kroiz, Gerson and Li, Feileen and Tao, Hudson and Srivastava, Ashay and Da Costa, Hevander and Gupta, Saloni and Rogers, Megan L. and Goncearenco, Inna and Sarli, Giuseppe and Galynker, Igor and Peskoff, Denis and Carpuat, Marine and White, Jules and Anadkat, Shyamal and Hoyle, Alexander and Resnik, Philip},
  title = {The Prompt Report: A Systematic Survey of Prompt Engineering Techniques},
  year = {2024},
  eprint = {2406.06608},
  archivePrefix = {arXiv},
  url = {https://arxiv.org/abs/2406.06608},
  doi = {10.48550/arXiv.2406.06608}
}

@misc{sahoo2024promptsurvey,
  author = {Sahoo, Pranab and Singh, Ayush Kumar and Saha, Sriparna and Jain, Vinija and Mondal, Samrat and Chadha, Aman},
  title = {A Systematic Survey of Prompt Engineering in Large Language Models: Techniques and Applications},
  year = {2024},
  eprint = {2402.07927},
  archivePrefix = {arXiv},
  url = {https://arxiv.org/abs/2402.07927}
}

@misc{geng2025jsonschemabench,
  title = {{JSONSchemaBench}: A Rigorous Benchmark of Structured Outputs for Language Models},
  author = {Geng, Saibo and Cooper, Hudson and Moskal, Micha{\l} and Jenkins, Samuel and Berman, Julian and Ranchin, Nathan and West, Robert and Horvitz, Eric and Nori, Harsha},
  year = {2025},
  eprint = {2501.10868},
  archivePrefix = {arXiv},
  primaryClass = {cs.CL}
}

@misc{wang2025slot,
  title = {{SLOT}: Structuring the Output of Large Language Models},
  author = {Wang, Darren Yow-Bang and Shen, Zhengyuan and Mishra, Soumya Smruti and Xu, Zhichao and Teng, Yifei and Ding, Haibo},
  year = {2025},
  eprint = {2505.04016},
  archivePrefix = {arXiv},
  primaryClass = {cs.CL}
}

@misc{ni2025llmbenchmarks,
  author = {Ni, Shiwen and Chen, Guhong and Li, Shuaimin and Chen, Xuanang and Li, Siyi and Wang, Bingli and Wang, Qiyao and Wang, Xingjian and Zhang, Yifan and Fan, Liyang and Li, Chengming and Xu, Ruifeng and Sun, Le and Yang, Min},
  title = {A Survey on Large Language Model Benchmarks},
  year = {2025},
  eprint = {2508.15361},
  archivePrefix = {arXiv},
  url = {https://arxiv.org/abs/2508.15361},
  doi = {10.48550/arXiv.2508.15361}
}

@inproceedings{sun2021chinesebert,
  author = {Sun, Zijun and Li, Xiaoya and Sun, Xiaofei and Meng, Yuxian and Ao, Xiang and He, Qing and Wu, Fei and Li, Jiwei},
  title = {ChineseBERT: Chinese Pretraining Enhanced by Glyph and Pinyin Information},
  booktitle = {Proceedings of ACL-IJCNLP 2021},
  pages = {2065--2075},
  year = {2021},
  url = {https://aclanthology.org/2021.acl-long.161/},
  doi = {10.18653/v1/2021.acl-long.161}
}

@inproceedings{huang2020multicriteria,
  author = {Huang, Kaiyu and Huang, Degen and Liu, Zhuang and Mo, Fengran},
  title = {A Joint Multiple Criteria Model in Transfer Learning for Cross-domain Chinese Word Segmentation},
  booktitle = {Proceedings of EMNLP 2020},
  pages = {3873--3882},
  year = {2020},
  url = {https://aclanthology.org/2020.emnlp-main.318/},
  doi = {10.18653/v1/2020.emnlp-main.318}
}

\newpage
\appendix
\section{Stage-level Loss Components}
\label{app:fn-dist}

Among the 3{,}495 strict false negatives, Stage~1 contributes
70.2\% (2{,}455 entries, almost entirely from the top-400k
frequency limit); Stage~4B contributes 13.4\% (470 entries);
Stages 2--4A combined contribute only 16.3\%. Detailed
component breakdown is in Table~\ref{tab:fn-dist}.

\begin{table*}[t]
\centering
\caption{False negative distribution by stage and component (total FN = 3,495)}
\label{tab:fn-dist}
\begin{tabular*}{\textwidth}{@{\extracolsep{\fill}}llrr@{}}
\toprule
Stage & Component & Count & \% of Total FN \\
\midrule
Stage 1 & length coverage limit & 191 & 5.5\% \\
Stage 1 & top-400k frequency limit & 2,264 & 64.8\% \\
Stage 2 & early pre-filter & 126 & 3.6\% \\
Stage 2 & XHC5 anchor & 42 & 1.2\% \\
Stage 2 & PMI loose & 350 & 10.0\% \\
Stage 3 & LLM FRAGMENT misclassification & 28 & 0.8\% \\
Stage 4A & rule false rejection & 24 & 0.7\% \\
Stage 4B & ENTITY (strict criterion) & 103 & 2.9\% \\
Stage 4B & NONE misclassification & 367 & 10.5\% \\
Total &  & 3,495 & 100.0\% \\
\bottomrule
\end{tabular*}
\end{table*}

\section{Length and Stage 4 Diagnostics}
\label{app:length-stage4}

\begin{table*}[t]
\centering
\setlength{\tabcolsep}{3pt}
\caption{Per-stage conditional recall by candidate length}
\label{tab:by-length}
\begin{tabular}{@{}>{\raggedright\arraybackslash}p{0.28\textwidth}>{\raggedright\arraybackslash}p{0.215\textwidth}>{\raggedright\arraybackslash}p{0.215\textwidth}>{\raggedright\arraybackslash}p{0.215\textwidth}@{}}
\toprule
Stage & 2-char (n=2,110) & 3-char (n=1,245) & 4-char (n=653) \\
\midrule
Stage 1: character n-gram & 56.87 [54.75, 58.97] & 24.90 [22.58, 27.38] & 35.83 [32.25, 39.59] \\
Stage 2: XHC5 + PMI & 58.42 [55.61, 61.17] & 95.16 [92.17, 97.05] & 98.29 [95.69, 99.33] \\
Stage 3: LLM well-formedness skill & 96.86 [95.29, 97.92] & 97.97 [95.63, 99.06] & 100.00 [98.36, 100.00] \\
Stage 4A: rule pre-filter & 98.53 [97.31, 99.20] & 97.92 [95.55, 99.05] & 96.52 [93.29, 98.23] \\
Stage 4B: LLM 3-way skill (strict NEO) & 65.62 [61.94, 69.12] & 59.01 [53.20, 64.58] & 44.14 [37.77, 50.72] \\
Pipeline strict recall & 20.81\% & 13.41\% & 15.01\% \\
\bottomrule
\end{tabular}

\vspace{0.8em}
\caption{Stage 4 internal classification on Hou (2023) entries reaching Stage 4 (n = 1,198)}
\label{tab:stage4-diag}
\begin{tabular}{lrr}
\toprule
Status & Count & Share \\
\midrule
Correct NEOLOGISM (strict TP) & 704 & 58.8\% \\
ENTITY (lexical innovation) & 103 & 8.6\% \\
Stage 4A rule false rejection & 24 & 2.0\% \\
Stage 4B NONE misclassification & 367 & 30.6\% \\
\bottomrule
\end{tabular}
\end{table*}


\end{document}